\documentclass[a4paper,12pt,onecolumn]{article}
\usepackage{setspace}
\usepackage{graphicx,graphics,times,amsmath,amssymb} 

\usepackage{t1enc}
\usepackage[latin1]{inputenc}

\begin{document}

\title{Fuzzy-Based Dialectical Non-Supervised Image Classification and Clustering\thanks{W. P. dos Santos (corresponding author) is with Departamento de Engenharia Elétrica, Universidade Federal de Campina Grande, Campina Grande, PB, Brazil, and with Escola Politécnica de Pernambuco, Universidade de Pernambuco, Recife, PE, Brazil (email: wellington.santos@ee.ufcg.edu.br); F. M. de Assis is with Departamento de Engenharia Elétrica, Universidade Federal de Campina Grande, Campina Grande, PB, Brazil (fmarcos@dee.ufcg.edu.br); Ricardo E. de Souza is with Departamento de Física, Universidade Federal de Pernambuco, Recife, PE, Brazil; Priscilla B. Mendes, Henrique S. S. Monteiro, and Havana D. Alves are with Escola Politécnica de Pernambuco, Universidade de Pernambuco, Recife, PE, Brazil.}
}

\author{Wellington P. dos Santos, Francisco M. de Assis, Ricardo E. de Souza,\\Priscilla B. Mendes, Henrique S. S. Monteiro, and Havana D. Alves}

\maketitle

\begin{abstract}
The materialist dialectical method is a philosophical investigative method to analyze aspects of reality. These aspects are viewed as complex processes composed by basic units named poles, which interact with each other. Dialectics has experienced considerable progress in the 19th century, with Hegel's dialectics and, in the 20th century, with the works of Marx, Engels, and Gramsci, in Philosophy and Economics. The movement of poles through their contradictions is viewed as a dynamic process with intertwined phases of evolution and revolutionary crisis. In order to build a computational process based on dialectics, the interaction between poles can be modeled using fuzzy membership functions. Based on this assumption, we introduce the Objective Dialectical Classifier (ODC), a non-supervised map for classification based on materialist dialectics and designed as an extension of fuzzy c-means classifier. As a case study, we used ODC to classify 181 magnetic resonance synthetic multispectral images composed by proton density, $T_1$- and $T_2$-weighted synthetic brain images. Comparing ODC to k-means, fuzzy c-means, and Kohonen's self-organized maps, concerning with image fidelity indexes as estimatives of quantization distortion, we proved that ODC can reach almost the same quantization performance as optimal non-supervised classifiers like Kohonen's self-organized maps.
\end{abstract}


\section{Introduction}

The dialectical conception of reality conceives all aspects of reality as complex processes composed by basic units called poles. Dialectics has its very roots in philosophical developments of the ancient civilizations of Greece, China and India, specially related to theories of Heraclite, Plato, and the philosophies of Confucianism, Buddhism, and Zen. As a universal method of analysis, dialectics has experienced considerable progress related to the development of German Philosophy in the 19th century, with Hegel's dialectics and the posterior works of Marx and Engels and, in the 20th century, the works of Gramsci and the Frankfurt School, in Philosophy and Economics. All those philosophers produced seminal works on the dynamics of contradictions in class-based societies, giving rise to the philosophical school of Historical Materialism \cite{marx1980_1,gramsci1992_1,gramsci1992_2,bobbio1990}.

The dialectical method of Historical Materialism was also proposed as a tool to study natural and social systems by regarding the dynamics of contradictions between their integrating poles. This movement of poles through their contradictions is viewed as a dynamic process with intertwined phases of \emph{evolution} and \emph{revolutionary crisis}. This was the principal inspiration to us to conceive a kind of evolutionary method for classification able to solve pattern recognition problems, mainly clustering and non-supervised classification.

Each of the most common paradigms of Computational Intelligence, namely neural networks, evolutionary computing, and culture-inspired algorithms, has its basis in a kind of theory intended to be of general application, but in fact very incomplete; e.g. the neural networks approach is based on a certain model of the brain; evolutionary computing is based on Darwin's theory; and cultural-inspired algorithms are based on the study of populations, such as those of ant colonies. Nevertheless, it is important to note that it is not necessarily the case, and indeed it may be impossible, that the theories an algorithm are based on have to be complete. In fact, it is utopy. For example, neural networks utilize a well-known incomplete model of the neurons, but useful to be applied for learning strategies. Furthermore, evolutionary algorithms are simplified mathematical models of natural evolution, but adequate to solve optimization problems. Consequently, there is a strong reason for investigating the use of Philosophy as a source of inspiration for developing computational intelligent methods and models to apply in several areas: dialectics and their features derived from the properties of unity, causality and conflict of opposites could be interesting to construct self-adaptative clustering methods where the number of clusters vary. These properties could be useful in areas such as pattern recognition \cite{santos2008a}.

The works of Thornley and Gibb discussed the application of the dialectical method to comprehend more clearly the paradoxical and conceptually contradictory discipline of information retrieval \cite{thornley2007}, while works of Rosser Jr. attempted to use several concepts of dialectics in nonlinear dynamics, making comparisons between aspects of Marx and Engel's dialectical method and concepts of Catastrophe Theory, Emergent Dynamics Complexity and Chaos Theory \cite{rosser2000}. Despite such theoretical efforts, there exist just a very few works concerned in composing mathematical approaches to rebuild the fundamental concepts of dialectics as mathematically useful tools to construct computational methods \cite{santos2008a}.

Santos \emph{et al.} introduced the Objective Dialectical Classifier (ODC), an evolutionary computational method that operates as a non-supervised self-organized map dedicated to classification \cite{santos2008a}. ODC is based on the dynamics of contradictions between dialectical poles. For classification, each class is considered as a dialectical pole. These poles interact with each other. This interaction is called pole struggle and is modeled using fuzzy membership functions. These poles are also affected by revolutionary crises, when some poles may disappear or be absorbed by other ones. New poles can emerge from periods of revolutionary crisis. The process of pole struggle and revolutionary crisis tends to a stable system, e.g. a system corresponding to the clustering of the original data, as our experiments are going to show.

In this work, as a case study, we used ODC to classify magnetic resonance (MR) synthetic multispectral images composed by proton density, $T_1$- and $T_2$-weighted synthetic images of 181 slices with 1~mm, spatial resolution\footnote{Dealing with MR, Positron Emission Tomography (PET), Single Photon Emission Computed Tomography (SPECT), and X-ray computerized tomography images, it is common to express spatial resolution using volume units.} of 1~mm$^3$, for a normal brain and a noiseless MR tomographic system without field inhomogeneities, amounting a total of 543 images, generated by the simulator BrainWeb \cite{brainweb1998}. Our principal target here is comparing ODC with other non-supervised classifiers, namely k-means, classical fuzzy c-means, and Kohonen's self-organized maps, concerned with image fidelity indexes as estimatives of quantization distortion.

This work is organized as follows: section \ref{sec:odm} exhibits general and specific definitions of ODC; section \ref{sec:methods} shows image fidelity expressions, parameters of non-supervised image classification methods, and synthetic brain MR images used in this work; quantitative and qualitative experimental results for image quantization are presented in section \ref{sec:results}, where in section \ref{sec:discussion} some discussion on experimental results is performed; finally, conclusions are presented in section \ref{sec:conclusion}.

\section{The Objective Dialectical Method} \label{sec:odm}

The Objective Dialectical Method (ODM) is an evolvable method designed to model dynamic systems and to perform tasks of classification, pattern recognition, intelligent search and optimization. It is based on the following features: a dialectical system is composed by several basic units, called poles. These poles interact with each other and are influenced by external conditions (inputs). These conditions are modeled as vectors of weights. The dialectical method is organized in phases of evolution (also called historical phases) and phases of revolutionary crisis. In phases of evolution, the weights of the poles are adjusted according to the inputs and the weights of the other poles, in an iterative process.

Each pole is associated to a determined measure of force, representing a performance evaluation. Such a measure of force can be both a degree of similarity, as it is common in clustering and pattern recognition applications, and the evaluation of an objective function in optimization problems. The stronger the pole, i.e. the better the performance of a determined pole, the greater the influence of this pole on the adjustment of the weights of the other poles according to its vectorial direction. At the end of each phase of evolution, the dialectical system is submitted to crises, i.e. the poles whose forces are under a determined threshold are eliminated; the similar poles (lower contradictions between each other) are fused; the most different poles (larger contradictions between each other) are used to generate new poles\footnote{It is important to notice that, in Dialectics, contradiction is a concept opposed to similarity: the greater the contradiction between two vectors, the smaller the similarity between them.}. After this process, some noise is added to the weights of the poles before a new historical phase starts. Therefore, important input parameters are: the number of historical phases, the duration of each historical phase, and the initial number of poles. The following steps present more details on the description of the dialectical algorithm:

\begin{enumerate}
	\item System inputs must be represented as a \emph{vector of conditions} representing the main features of the problem;
	\item The user has to provide the initial parameters: the initial number of \emph{poles} (corresponding to clusters or classes in problems of unsupervised classification) that compose the system, the \emph{number of historical phases}, and \emph{duration of each historical phase}. The number of historical phases and their duration can also be randomly defined, depending on the application;
	\item Each dialectical pole is associated with: (i) a \emph{vector of weights}, with the same size as the vector of conditions; (ii) a \emph{similarity function}, and (iii) a \emph{measure of force}. The initial vectors of weights can be randomly defined or chosen from the set of vectors of conditions;
	\item The historical phases consist of two stages:
	\begin{enumerate}
		\item \textbf{Evolution:} The poles compete with each other as we described above. In Dialectics, this competition process is also called \emph{pole struggle}. A similarity function associated with each pole is evaluated and, given a vector of conditions, the winner pole, which is the pole with greatest degree of similarity according to a determined vector of conditions, has its parameters (weights and measure of force) incremented. This process continues until the end of the historical phase is reached;
		\item \textbf{Revolutionary crisis:} This starts at the end of the historical phase. At this point the following steps are performed:
		\begin{enumerate}
			\item The measures of force are compared, i.e. all the poles with a measure of force less than a minimum force are marked;
			\item The contradictions between the integrating poles are also evaluated. This evaluation is performed using the similarity functions as described in the next paragraphs. If a contradiction between two poles is less than a given minimum contradiction, one of the two poles is selected or marked as such. Here the minimum contradiction plays the role of a threshold;
			\item From the evaluated contradictions computed in the previous items, the overall maximum contradiction is calculated. This is \emph{the main contradiction} of the system. From the pair of poles involved in the main contradiction, a new pole is generated, i.e. a synthesis of previous pairs of poles, whose vectors of weights are calculated from the vector of weights of such pairs;
			\item All the marked poles are eliminated and a new set of poles is generated;
			\item The vectors of weights of all poles of the new set of poles are randomly modified, representing the impact of the revolutionary crisis on both the survivors and the new poles.
		\end{enumerate}
	\end{enumerate}
\end{enumerate}

From the previous general definition it is possible to generate several specific definitions. A proposal inspired in fuzzy c-means maps based on the Principle of Maximum Entropy \cite{chen2009} is the following:

Let $\mathbf{x}=(x_1, x_2, \dots, x_n)^T$ be a vector with $n$ conditions influencing the dialectical system with $n_C(t)$ poles at instant $t$, since $\Omega(t)=\{C_1, C_2, \dots, C_{n_C(t)}\}$ is the set of poles of a determined system, where each pole $C_k$ is associated to a fuzzy membership function, i.e. a similarity function $g_k:\mathbb{R}\rightarrow [0,1]$ and a vector of weights $\mathbf{w}_k$, defined as following:
\begin{equation}
  g_k(\mathbf{x})=\frac{\exp(-\frac{1}{n_C(t)}||\mathbf{x}-\mathbf{w}_k(t)||)}{\sum_{l=1}^{n_C(t)} \exp(-\frac{1}{n_C(t)}||\mathbf{x}-\mathbf{w}_l(t)||)},
\end{equation}
\begin{equation}
  \mathbf{w}_k=(w_{k,1}, w_{k,2}, \dots, w_{k,n})^T,
\end{equation}
for $1\leq k\leq n_C(t)$, where this way of implementation was chosen to model the influence of the winner pole over the other poles considering force and similarity, as we described above. The index $t$ indicates the iteration (or time). Notice that $g_k(\mathbf{x})$ expresses the similarity between the vector of conditions $\mathbf{x}$ and the vector of weights of the $k$-th pole. The expression of these membership functions is based on Gibb's distribution \cite{jaynes1957a}, as it was proved by Chen and Zhao, and obtained by the application of the Principle of Maximum Entropy to modify the classical fuzzy c-means clustering method, changing probabilities by membership function values, to improve the ability of the algorithm to find new cluster centroids by the maximization of fuzzy-based entropy \cite{chen2009}.

The similarity functions are also used to evaluate the contradictions between the poles: $g_i(\mathbf{w}_j)=g_j(\mathbf{w}_i)$ indicate the degree of similarity between poles $C_i$ and $C_j$; alternatively, the contradiction between poles $C_i$ and $C_j$ are given by $\delta_{i,j}=\delta_{j,i}$, where $\delta_{i,j}=1-g_i(\mathbf{w}_j)$ and $1\leq i,j\leq n_C(t)$.

Let $n_P$ be the maximum number of historical phases, $n_H(t)$ be the duration of each historical phase, and $\eta(t)$ be the step size, where $0<\eta(t)<1$, the algorithms runs until a determined number of poles is reached or a determined cost function is minimized. The step size $\eta(t)$ develops herein this work the same role as the learning rate in neural networks.

The stage of \emph{evolution} or \emph{pole struggle} can be implemented as follows:
\begin{equation}
  w_{i,j}(t+1)=\left\{
  \begin{array} {ll}
      {w_{i,j}(t)+\Delta w_{i,j}(t),} & {i=k(t)}\\
      {w_{i,j}(t),} & {i\neq k(t)}
    \end{array}
  \right.,
\end{equation}
\begin{equation}
  \Delta w_{i,j}(t)=\eta(t)g_i^2(\mathbf{x})(x_j(t)-w_{i,j}(t)),
\end{equation}
\begin{equation}
  f_{i}(t+1)=\left\{
  \begin{array} {ll}
      {f_{i}(t)+1,} & {i=k(t)}\\
      {f_{i}(t),} & {i\neq k(t)}
    \end{array}
  \right.,
\end{equation}
for $1\leq i\leq n_C(t)$ and $1\leq j\leq n$, where $f_i$ is the measure of force associated to pole $C_i$, and $k$ is the index of the winner pole:
\begin{equation}
  k(t)=\arg \max\{ g_1(\mathbf{x}), g_2(\mathbf{x}), \dots, g_{n_C(t)}(\mathbf{x}) \}.
\end{equation}

In the stage of \emph{revolutionary crisis}, a determined \emph{binary vector of marks} is defined as following:
$$
\mathbf{m}=(m_1, m_2, \dots, m_{n_C(t)})^T,
$$
where $m_i=1$ when pole $C_i$ is absorbed by other pole or simply eliminated, and $m_i=0$ on the contrary, for $1\leq i\leq n_C(t)$. This vector is initially null. It is important to notice that such a vector is represented just to help our explanation, because its use is not computationally efficient, once marked poles can be eliminated without such a vector definition.

In the stage of revolutionary crisis, the measures of force are evaluated first. These measures are normalized as following:
\begin{equation}
  \bar{f}_i(t)=\frac{f_i(t)}{\max\{f_l(t)\}_{l=1}^{n_C(t)}},
\end{equation}
for $1\leq i\leq n_C(t)$. If a determined normalized measure of force is less than the \emph{minimum measure of force} necessary for the associated pole to survive after pole struggle, $f_{\min}$, that is, $\bar{f}_i(t)<f_{\min}$, we will have $m_i=1$.

After evaluating forces of each pole, the contradictions among them are evaluated. The contradiction between poles $C_i$ and $C_j$, represented by $\delta_{i,j}$, is determined by $\delta_{i,j}=1-g_i(\mathbf{w}_j)$, where $1\leq i,j\leq n_C(t)$. If the contradiction is less than the \emph{minimum contradiction}, $\delta_{\min}$, that is, $\delta_{i,j}<\delta_{\min}$, poles $C_i$ and $C_j$ are considered effectively the same pole, and one of the poles is marked. In case pole $C_i$ is marked, $m_i=1$. Obviously, these expressions are also dependant on $t$. However, we decided to ommit this fact, to get a simpler expression.

Once contradictions are evaluated, the partial set of integrating poles of the new dialectical system is generated, $\Omega'(t)$, as following:
\begin{equation}
  m_i(t)=0 \Leftrightarrow C_i\in \Omega'(t),
\end{equation}
for $1\leq i\leq n_C(t)$.

The search for the \emph{main contradictions}, or \emph{principal contradictions}, is performed by putting the set of the contradictions $\delta_{i,j}$ (notice $\delta_{i,j}=\delta_{j,i}$, for $i\neq j$ and $m_i=m_j=0$, and then taking the $p\geq 1$ greatest values. From the poles involved in principal contradictions we generate new poles $C_k$, where $k>n_C(t)$. This process is closely related to the dialectical concept of praxis regarding the generation of new poles.

The vector of weights associated to the new poles $C_k$, $\mathbf{w}_k$, can be calculated from the following process, inspired in crossover operators of genetic algorithms:
\begin{equation}
  w_{k,r}(t+1)=\left\{
  \begin{array}{ll}
    {w_{p,r}(t+1),} & {r~\textnormal{mod}~2=1}\\
    {w_{q,r}(t+1),} & {r~\textnormal{mod}~2=0}
  \end{array} \right.,
\end{equation}
where $1\leq r\leq n$, while $\mathbf{w}_p$ and $\mathbf{w}_q$ are the vectors of weights of poles $C_p$ and $C_q$, respectively, involved in principal contradictions, i.e. $\delta_{p,q}=\max\{\delta_{i,j}\}$, for $1\leq i,j\leq n_C(t)$ and $i\neq j$. These new poles, $C_k$, compose the \emph{set of new poles}, $\Omega''(t)$. Therefore, the set of new poles, $\Omega'''(t)$, is obtained from the following expression:
\begin{equation}
  \Omega'''(t)=\Omega'(t)\cup\Omega''(t).
\end{equation}

The qualitative influence of the revolutionary crisis over new and reminiscent poles is modeled by the \emph{function of crisis}, $\chi(t)$, defined by:
\begin{equation}
  \chi(t)=\chi_{\max}(t)G(0,1),
\end{equation}
where $\chi_{\max}(t)$ is a parameter called \emph{maximum crisis}, and $G(0,1)$ is a random number distributed according to the distribution of Gauss, with expectance 0 and variance 1. Therefore, the new set of poles is
\begin{equation}
  \Omega(t+1)=\{ C_1(t+1), C_2(t+1), \dots, C_{n_C(t+1)}(t+1) \},
\end{equation}
where
\begin{equation}
  C_k(t+1)=C_k(t)\in \Omega'''(t),
\end{equation}
and
\begin{equation}
  w_{k,i}(t+1)=w_{k,i}(t)+\chi(t),
\end{equation}
for $1\leq k\leq n_C(t+1)$ and $1\leq i\leq n$. This function of crisis is just a mathematical way to model the influence of crisis in dialectical transitions by adding determined levels of random noise. The stage of revolutionary crisis and, consequently, the historical phase, come to an end. A new historical phase begins, continuing while the maximum number of historical phases is not reached.

\section{Objective Dialectical Classifiers}

Objective Dialectical Classifiers are an adaptation of ODM to tasks of classification. This means that the feature vectors are mounted and considered as vectors of conditions. Specifically, once they are applied to the inputs of the dialectical system, their coordinates will affect the dynamics of the contradictions among the integrating dialectical poles. Hence, the integrating poles model the recognized classes at the task of non-supervised classification \cite{santos2008a}. In order to guarantee the convergence of the dialectical classifier, we have removed the operator of pole generation, present at the revolutionary crises. Therefore we can guarantee that the number of poles at the end of the iterations is minor or equal to the initial number of poles \cite{santos2008a,santos2009a,santos2009c,santos2009d,santos2009e,santos2009f,santos2009g,santos2009h}.

Consequently, an objective dialectical classifier is in fact an evolvable non-supervised classifier where, instead of supposing a predetermined number of classes, we can set an initial number of classes (i.e. dialectical poles) and, as the historical phases happen as a result of pole struggles and revolutionary crises, classes are eliminated or absorbed, whilst new classes are generated. After the end of the training process, the dialectical system presents a number of statistically significant classes present in the training set and, therefore, a feasible classifier associated to the final state of the dialectical system \cite{santos2008a,santos2008b}.

Once the training process is complete, objective dialectical classifier behavior occurs in the same way as any non-supervised classification method. This is clear if we analyze the training process when $n_P=n_H=1$. This transforms the dialectical classifier into the fuzzy c-means classifier based on the Principle of Maximum Entropy \cite{chen2009}.

The non-supervised classification process is performed as the following described manner: given a set of input conditions
\begin{equation}
  \mathbf{x}=(x_1, x_2, \dots, x_n)^T,
\end{equation}
if the learning process reached stabilization with
\begin{equation}
  \Omega=\{C_1, C_2, \dots, C_{n_C}\},
\end{equation}
the classification rule is defined as following:
\begin{equation}
  k=\arg\max\{g_1(\mathbf{x}), g_2(\mathbf{x}), \dots, g_{n_C}(\mathbf{x})\}\Rightarrow \mathbf{x}\in C_k.
\end{equation}

\section{Materials and Methods} \label{sec:methods}

\subsection{MR Images}

In this work we adopted the following case study: we used magnetic resonance (MR) synthetic multispectral images composed by proton density, $T_1$- and $T_2$-weighted synthetic sagital images of 181 slices with 1~mm, resolution of 1~mm$^3$, for a normal brain and a noiseless MR tomographic system without field inhomogeneities, amounting a total of 543 images, generated by MR image simulator BrainWeb \cite{brainweb1999,brainweb1998}. These images can have a maximum amount of 13 anatomical elements. Therefore, the number of classes present in each image can reach 13 classes. Consequently, each class is associated to a determined output of the classifiers we used to perform this study.

\begin{figure}
	\centering
		\includegraphics[width=0.35\textwidth]{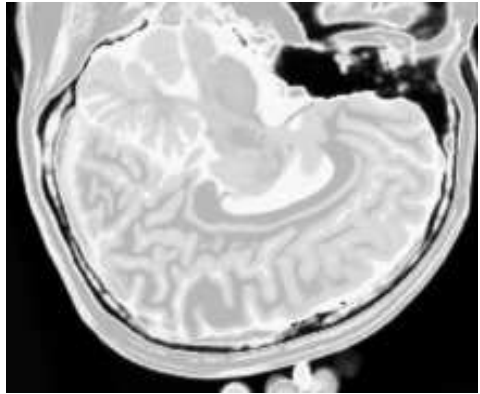}
	\caption{PD-weighted MR image of the 97th slice}
	\label{fig:97_normal_pn0_rf0_pd}
\end{figure}

\begin{figure}
	\centering
		\includegraphics[width=0.35\textwidth]{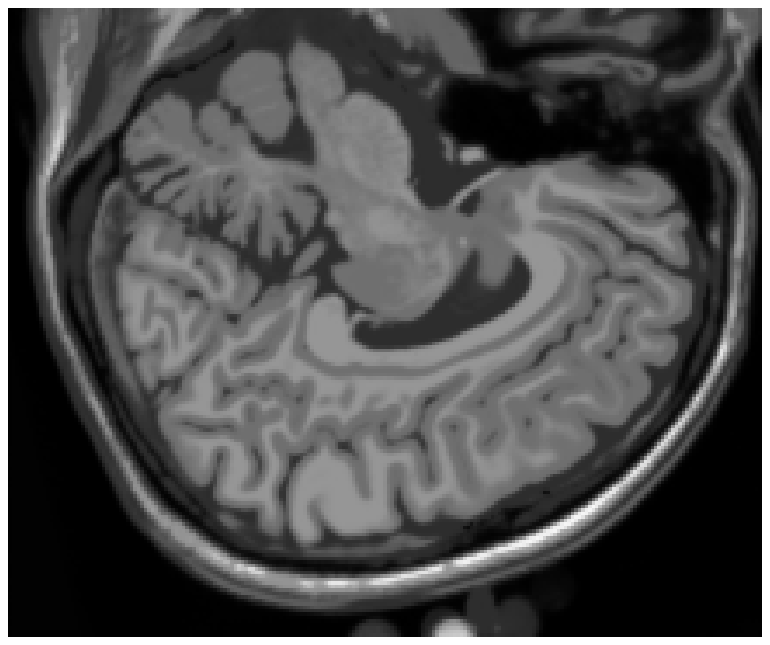}
	\caption{$T_1$-weighted MR image of the 97th slice}
	\label{fig:97_normal_pn0_rf0_t1}
\end{figure}

\begin{figure}
	\centering
		\includegraphics[width=0.35\textwidth]{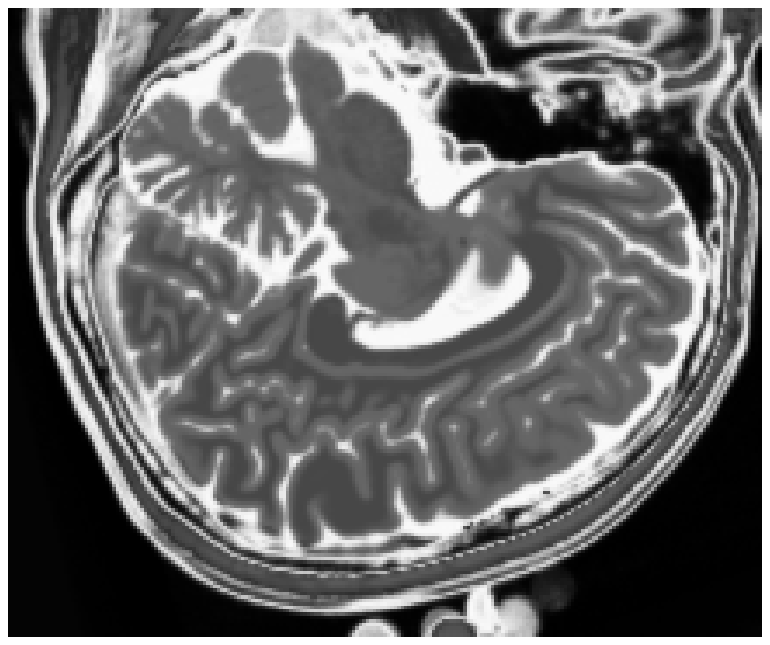}
	\caption{$T_2$-weighted MR image of the 97th slice}
	\label{fig:97_normal_pn0_rf0_t2}
\end{figure}

\begin{figure}
	\centering
		\includegraphics[width=0.35\textwidth]{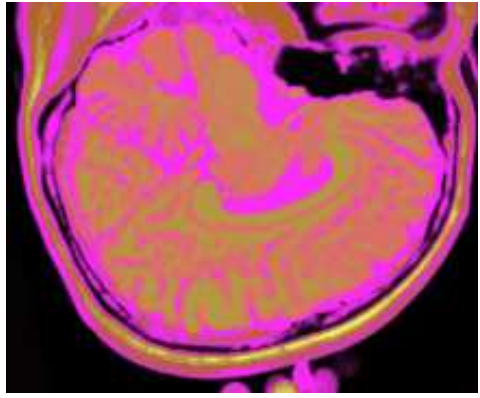}
	\caption{R0-G1-B2 colored composition of PD-, $T_1$-, and $T_2$-weighted MR images of the 97th slice}
	\label{fig:97_normal_pn0_rf0}
\end{figure}

Figures \ref{fig:97_normal_pn0_rf0_pd} (band 0), \ref{fig:97_normal_pn0_rf0_t1} (band 1) and \ref{fig:97_normal_pn0_rf0_t2} (band 2) show PD- (proton density), $T_1$- and $T_2$-weighted MR images of the 97th slice, while figure \ref{fig:97_normal_pn0_rf0} shows the R0-G1-B2 colored composition of the same slice.

\subsection{Image Fidelity Indexes}

The distortion error for all non-supervised classification methods was indirectly measured using the following global pixel-by-pixel fidelity indexes: the maximum error (ME), the mean absolute error (MAE), the mean square error (MSE), the root mean square error (RMSE), and the peak signal to noise ratio (PSNR), described as following \cite{pedrini2008}:
\begin{equation}
  \epsilon_\textnormal{ME}=\max\{||f(\textbf{u})-f'(\textbf{u})||\}_{\textbf{u}\in S},
\end{equation}
\begin{equation}
  \epsilon_\textnormal{MAE}=\frac{1}{\#S} \sum_{\textbf{u}\in S} ||f(\textbf{u})-f'(\textbf{u})||, 
\end{equation}
\begin{equation}
  \epsilon_\textnormal{MSE}=\frac{1}{\#S} \sum_{\textbf{u}\in S} ||f(\textbf{u})-f'(\textbf{u})||^2, 
\end{equation}
\begin{equation}
  \epsilon_\textnormal{RMSE}=\sqrt{\epsilon_\textnormal{MSE}}, 
\end{equation}
\begin{equation}
  \epsilon_\textnormal{PSNR}=20\log_{10}\frac{L_{\max}}{\epsilon_\textnormal{RMSE}}, 
\end{equation}
for $W= \{0,1,\dots,L_{\max}\}$, considering a $n$-band multispectral image $f:S\rightarrow W^n$ and its reference image $f':S\rightarrow W^n$. For normalized multispectral images, $f:S\rightarrow [0,1]^n$, we have $L_{\max}=1$.

Obviously, there are other important and useful image fidelity indexes, like Wang and Bovik's index \cite{wang2002}. However, in this work we are interested in using image fidelity indexes as an indirect measure of the mean quantization distortion associated to non-supervised classification methods. Therefore, we are just interested in indexes based on pixel-by-pixel differences, once Wang and Bovik's index and other similar indexes are focused in comparisons based on image global statistics \cite{wang2002,pedrini2008}. This justifies our preference by simple and classical pixel-by-pixel fidelity indexes.

\subsection{Non-Supervised Image Classification Methods}

The synthetic multispectral images obtained by colored compositions R0-G1-B2 were classified using the following methods, also used to evaluate vector quantization performance:
\begin{enumerate}
  \item \emph{Kohonen self-organized map classifier (KO)}: 3 inputs, 13 outputs, maximum of 200 iterations, initial learning rate $\eta_0=0.1$, Gaussian function of distance;
  \item \emph{Fuzzy c-means classifier (CM)}: 3 inputs, 13 outputs, maximum of 200 iterations, initial learning rate $\eta_0=0.1$;
  \item \emph{K-means classifier (KM)}: 3 inputs, 13 outputs, maximum of 200 iterations, initial learning rate $\eta_0=0.1$.
	\item \emph{Objective dialectical classifier (ODC)}: 14 initial poles, 2 historical phases of 150 iterations each phase, initial historical step $\eta_0=0.1$, minimum force of $5\%$, minimum contradiction of $1\%$, maximum contradiction of $98\%$, maximum crisis of $35\%$, until 12 poles. After all historical phases, the training process was finished with 13 poles.
\end{enumerate}

\section{Experimental Results} \label{sec:results}

\begin{figure}
	\centering
		\includegraphics[width=0.35\textwidth]{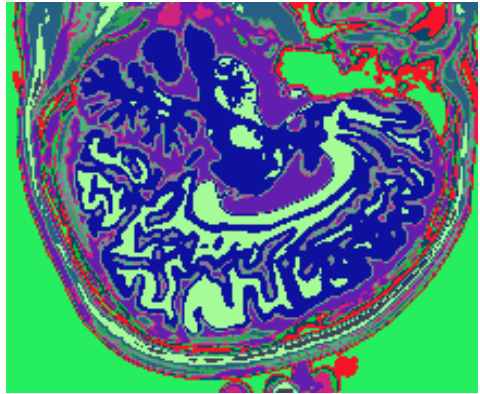}
	\caption{Classification results of the 97th slice using KO method}
	\label{fig:97_class_normal_pn0_rf0_KO}
\end{figure}

\begin{figure}
	\centering
		\includegraphics[width=0.35\textwidth]{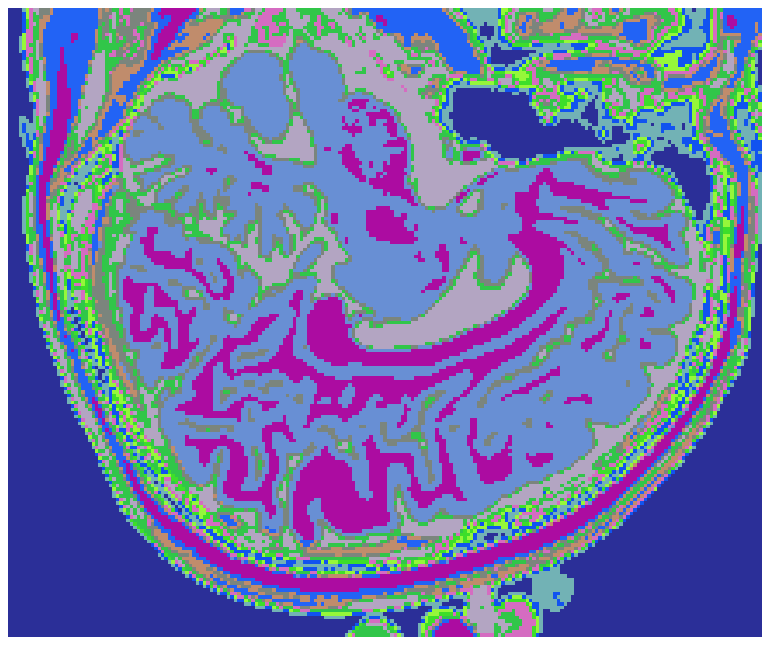}
	\caption{Classification results of the 97th slice using CM method}
	\label{fig:97_class_normal_pn0_rf0_CM}
\end{figure}

\begin{figure}
	\centering
		\includegraphics[width=0.35\textwidth]{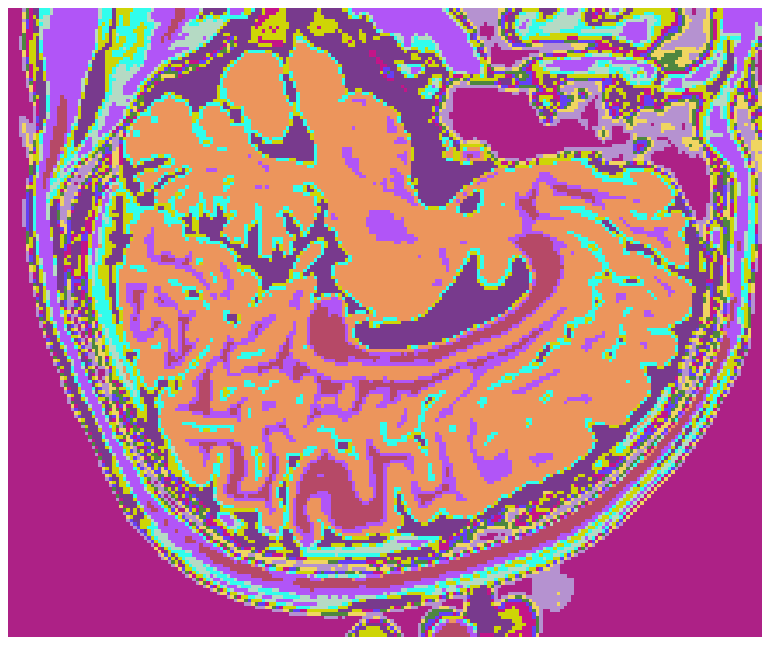}
	\caption{Classification results of the 97th slice using KM method}
	\label{fig:97_class_normal_pn0_rf0_KM}
\end{figure}

\begin{figure}
	\centering
		\includegraphics[width=0.35\textwidth]{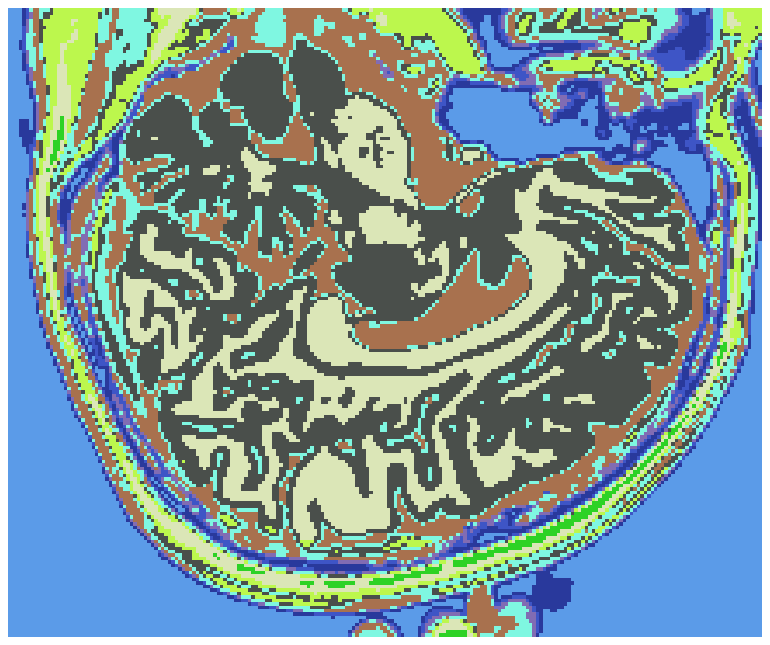}
	\caption{Classification results of the 97th slice using ODC method}
	\label{fig:97_class_normal_pn0_rf0_DLTPME}
\end{figure}

\begin{figure}
	\centering
		\includegraphics[width=0.35\textwidth]{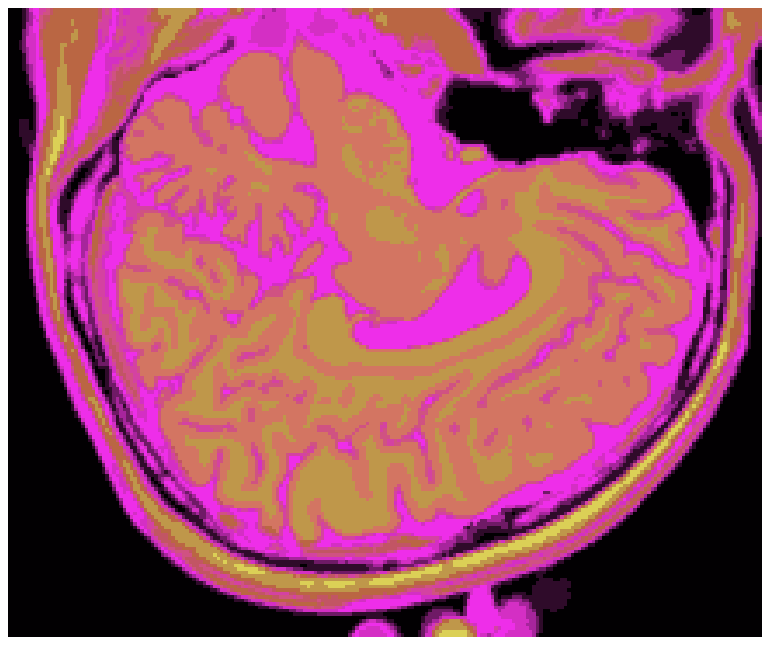}
	\caption{Quantization results of the 97th slice using KO method}
	\label{fig:97_quant_normal_pn0_rf0_KO}
\end{figure}

\begin{figure}
	\centering
		\includegraphics[width=0.35\textwidth]{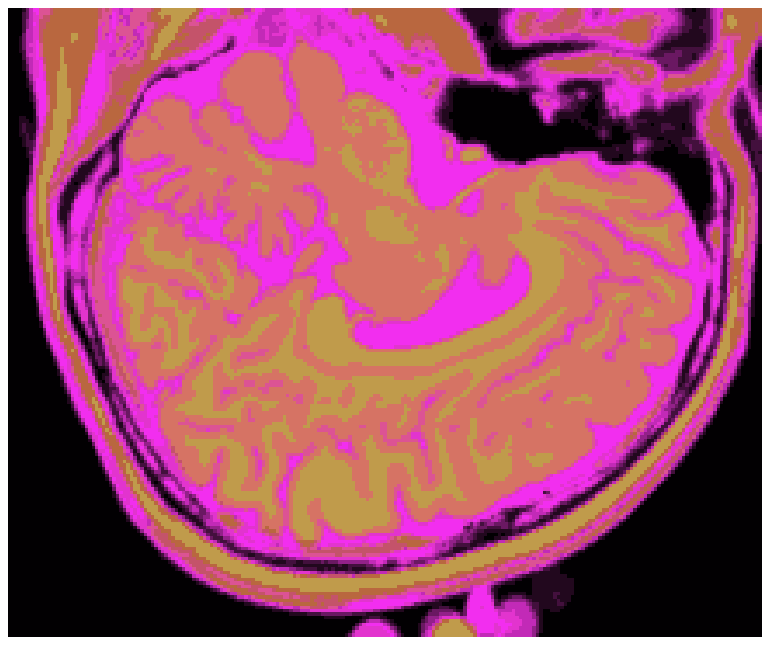}
	\caption{Quantization results of the 97th slice using CM method}
	\label{fig:97_quant_normal_pn0_rf0_CM}
\end{figure}

\begin{figure}
	\centering
		\includegraphics[width=0.35\textwidth]{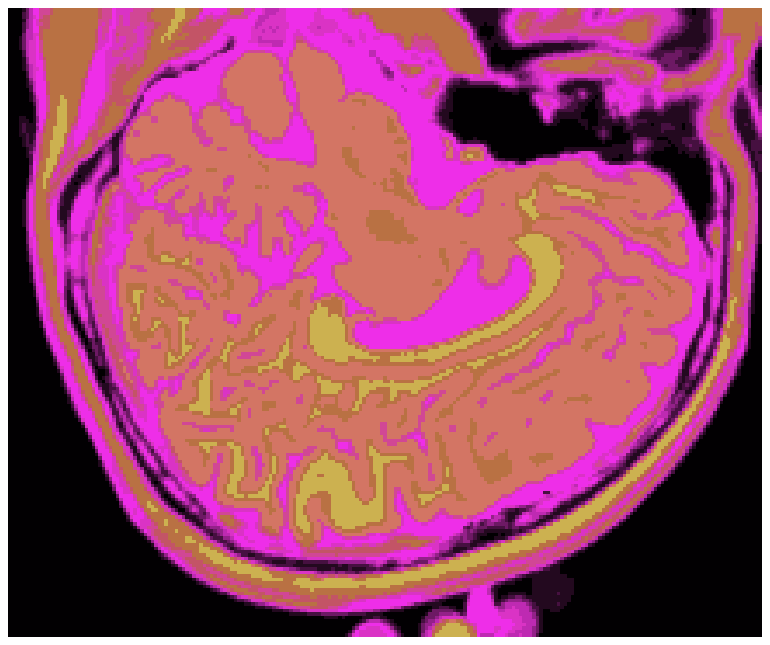}
	\caption{Quantization results of the 97th slice using KM method}
	\label{fig:97_quant_normal_pn0_rf0_KM}
\end{figure}

\begin{figure}
	\centering
		\includegraphics[width=0.35\textwidth]{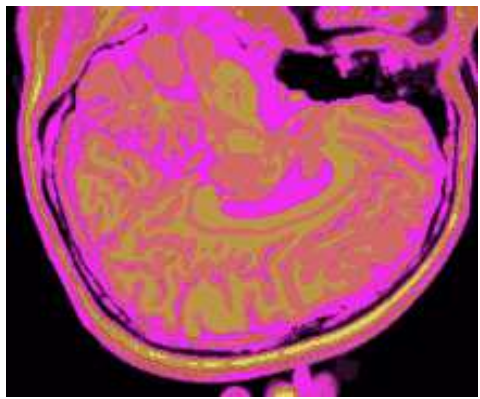}
	\caption{Quantization results of the 97th slice using ODC method}
	\label{fig:97_quant_normal_pn0_rf0_DLTPME}
\end{figure}

Figures \ref{fig:97_class_normal_pn0_rf0_KO}, \ref{fig:97_class_normal_pn0_rf0_CM}, \ref{fig:97_class_normal_pn0_rf0_KM} and \ref{fig:97_class_normal_pn0_rf0_DLTPME} show classification results, whilst figures \ref{fig:97_quant_normal_pn0_rf0_KO}, \ref{fig:97_quant_normal_pn0_rf0_CM}, \ref{fig:97_quant_normal_pn0_rf0_KM} and \ref{fig:97_quant_normal_pn0_rf0_DLTPME} exhibit quantization results for the image of the 97th slice, figure \ref{fig:97_normal_pn0_rf0}, using methods KO, CM, KM and ODC, respectively. Image quantization is the procedure of constraining a determined image from its complete set of pixels to a smaller set of vectors with same dimensions feasible to represent the original image with a smaller gamute, according to a given fidelity measure. Herein this work we built quantization images just by changing the original pixels for the vectors of weights related to the classification results, i.e. the centroids of the unsupervised classification methods. 

Table \ref{tab:quant} shows the results of the evaluation of the non-supervised classification methods with respect to vector quantization, using fidelity indexes $\epsilon_\textnormal{ME}$, $\epsilon_\textnormal{MAE}$, $\epsilon_\textnormal{MSE}$, $\epsilon_\textnormal{RMSE}$ and $\epsilon_\textnormal{PSNR}$, considering all 181 slices with 3 bands (DP, $T_1$ and $T_2$), for KO, CM, KM and ODC methods.

Table \ref{tab:individual_sim} shows one-by-one comparisons of pairs of sample mean and mean deviation of $\epsilon_\textnormal{ME}$, $\epsilon_\textnormal{MAE}$, $\epsilon_\textnormal{MSE}$, $\epsilon_\textnormal{RMSE}$ and $\epsilon_\textnormal{PSNR}$ for methods KO, CM, KM and ODC, according to the null-hypothesis test F, whilst table \ref{tab:global_sim} shows global comparisons between ODC method and KO, CM and KM methods, using test $\chi^2$ to measure the statistical adherence of results generated by ODC and results generated by the others methods; the sequences of observed values were build using the sample mean and the mean deviation of $\epsilon_\textnormal{ME}$, $\epsilon_\textnormal{MAE}$, $\epsilon_\textnormal{RMSE}$ and $\epsilon_\textnormal{PSNR}$. Both tests were performed taking into account 95\% of confidence.

\begin{table}[htbp]
	\centering
	\footnotesize
		\begin{tabular} {c|c|c|c|c}
			{} & {KO} & {CM} & {KM} & {ODC}\\
			\hline
			{$\epsilon_\textnormal{ME}$} & {$50\pm 5$} & {$106\pm 15$} & {$84\pm 22$} & {$50\pm 3$}\\
			{$\epsilon_\textnormal{MAE}$} & {$10\pm 3$} & {$10\pm 2$} & {$13\pm 4$} & {$12\pm 2$}\\
			{$\epsilon_\textnormal{MSE}$} & {$186\pm 52$} & {$258\pm 81$} & {$347\pm 261$} & {$249\pm 60$}\\
			{$\epsilon_\textnormal{RMSE}$} & {$13\pm 2$} & {$16\pm 31$} & {$18\pm 5$} & {$15\pm 2$}\\
			{$\epsilon_\textnormal{PSNR}$} & {$26\pm 2$} & {$24\pm 2$} & {$24\pm 3$} & {$25\pm 1$}\\
			\hline			
		\end{tabular}
	\caption{Quantization results concerning image fidelity indexes}
	\label{tab:quant}
\end{table}

\begin{table}[htbp]
	\centering
	\footnotesize
		\begin{tabular} {c|c|c|c|c}
			{} & {$\mu(\epsilon_\textnormal{ME})$} & {$\mu(\epsilon_\textnormal{MAE})$} & {$\mu(\epsilon_\textnormal{RMSE})$} & {$\mu(\epsilon_\textnormal{PSNR})$}\\
			\hline
			{ODC-KO} & {$0.99$} & {$0.86$} & {$0.89$} & {$1.00$}\\
			{ODC-CM} & {$0.61$} & {$0.86$} & {$0.89$} & {$1.00$}\\
			{ODC-KM} & {$0.83$} & {$0.86$} & {$0.89$} & {$1.00$}\\
			\hline			
		\end{tabular}
	\caption{Degrees of similarity of experimental results of each classification method according to null-hypothesis test F, for each similarity index}
	\label{tab:individual_sim}
\end{table}

\begin{table}[htbp]
	\centering
	\footnotesize
		\begin{tabular} {c|c|c}
			{ODC-KO} & {ODC-CM} & {ODC-KM}\\
			\hline
			{$0.96$} & {$0.00$} & {$0.00$}\\
			\hline			
		\end{tabular}
	\caption{Degrees of global similarity of experimental results of each classification method according to test $\chi^2$}
	\label{tab:global_sim}
\end{table}

\section{Discussion} \label{sec:discussion}

Table \ref{tab:quant} shows statistics of the fidelity indexes measured for each 3-band slice of the studied synthetic brain volume. From these results, regarding just sample means, we can see that ODC and KO got apparently the best results for $\epsilon_\textnormal{ME}$, $\epsilon_\textnormal{MAE}$ and $\epsilon_\textnormal{RMSE}$, respectively 46, 10 and 169 (ODC), against 50, 10, 186 (KO), 106, 10, 258 (CM), and 84, 13 and 347 (KM). However, concerning just sample means again, it is possible to see that these results were almost identical to the results obtained by KO and CM, considering $\epsilon_\textnormal{RMSE}$ and $\epsilon_\textnormal{PSNR}$, respectively 15 and 25 (ODC), 13 and 26 (KO), against 16 and 24 (CM), and 18 and 24 (KM). Such results point to an apparent little advantage of Kohonen's SOM classifier, once it is an optimal method taking into account vector quantization \cite{haykin2001}.

Herein this work we decided to use the null-hypothesis test F with 95\% of confidence, to make comparisons considering the sample means and the respective mean deviations of fidelity indexes. Table \ref{tab:individual_sim} shows these results, where we can see that, according to test F, there is a degree of similarity of 0.99 between ODC and KO, regarding $\epsilon_\textnormal{ME}$, against 0.61 and 0.83 for CM and KM, respectively. However, considering the other fidelity indexes, ODC is apparently very similar to the others: similarities of 0.86, 0.89 and 1.00, regarding indexes MAE, RMSE and PSNR, respectively, which means that ODC is practically equal to the other three methods, concercing PSNR. These results show that, concerning also table \ref{tab:quant} and one-by-one comparisons, although KO is obviously superior to the other methods, ODC method is reasonably close to KO.

We also decided to employ the adherence test $\chi^2$ with 95\% of confidence to get a global evaluation of ODC related to KO, CM and KM. Table \ref{tab:global_sim} shows the results of the application of test $\chi^2$ to the following pairs of sequences: ODC-KO, ODC-CM and ODC-KM. These sequences are composed by sample means of fidelity indexes $\epsilon_\textnormal{ME}$, $\epsilon_\textnormal{MAE}$, $\epsilon_\textnormal{RMSE}$ and $\epsilon_\textnormal{PSNR}$, and their respective mean deviations. The test results show that ODC results are similar to KO's, with similarity of 0.96.

\section{Conclusion} \label{sec:conclusion}

From the results we presented above we could conclude that the objective dialectical classifier is a good clustering method concerning quantization distorsion, able to get results almost as good as those results obtained by the optimal clustering method based on Kohonen self-organized maps. Table \ref{tab:quant} shows that KO is slightly better than ODC. However, we claim that ODC is a completely new algorithmic concept, having still a lot of unexplored design options, which could lead us to generate other forms and implementations for the dialectical method feasible to be used in other applications of pattern recognition and clustering. As future works, we propose the elaboration of new optimization and supervised classification-based methods using the dialectical analogies we presented in this work.

Therefore we can perceive that it is possible to construct feasible non-supervised classification and clustering methods based on philosophical heuristic and mimics approaches, taking into account more complex models than those biologically-inspired models commonly used in Computational Intelligence: models based on Philosophy and its investigative structural methods, refined by the use of the Principle of Maximum Entropy and the power of fuzzy membership functions to model non-probabilistic uncertainty.

\bibliographystyle{unsrt}
\bibliography{arq_bib}

\end{document}